\documentclass{article}

\usepackage{arxiv}

\usepackage[utf8]{inputenc} 
\usepackage[T1]{fontenc}    
\usepackage{hyperref}       
\usepackage{url}            
\usepackage{booktabs}       
\usepackage{amsfonts}       
\usepackage{nicefrac}       
\usepackage{microtype}      
\usepackage{lipsum}
\usepackage{graphicx}
\graphicspath{ {./images/} }
\usepackage{amsmath}
\usepackage{animate}

\title{A Multi-Branched Radial Basis Network Approach to Predicting Complex Chaotic Behaviours }

\author{
Aarush Sinha\\
aarush.sinha@gmail.com
}

\begin{document}
\maketitle
\begin{abstract}
    In this study, we propose a multi branched network approach to predict the dynamics of a physics attractor characterized by intricate and chaotic behavior. We introduce a unique neural network architecture comprised of Radial Basis Function (RBF) layers combined with an attention mechanism designed to effectively capture nonlinear inter-dependencies inherent in the attractor's temporal evolution. Our results demonstrate successful prediction of the attractor's trajectory across 100 predictions made using a real-world dataset of 36,700 time-series observations encompassing approximately 28 minutes of activity. To further illustrate the performance of our proposed technique, we provide comprehensive visualizations depicting the attractor's original and predicted behaviors alongside quantitative measures comparing observed versus estimated outcomes. Overall, this work showcases the potential of advanced machine learning algorithms in elucidating hidden structures in complex physical systems while offering practical applications in various domains requiring accurate short-term forecasting capabilities.
\end{abstract}


\section{Introduction}
In traditional mathematics, a radial basis function is a function that is based on the distance between the input and a specified point, such as the origin or a center point. A radial function is any function that meets this property \cite{rbf}. \\
A radial function is a function $\varphi:[0,\infty) \to \mathbb{R}$. When paired with a metric on a vector space $\|\cdot\|:V \to [0,\infty)$ a function $\varphi_{\mathbf{c}} = \varphi(\|\mathbf{x}-\mathbf{c}\|)$ is said to be a radial kernel centered at $\mathbf{c}$. A Radial function and the associated radial kernels are said to be radial basis functions if, for any set of nodes $\{\mathbf{x}_k\}_{k=1}^n$:
\begin{itemize}
    \item The kernels $\varphi_{\mathbf{x}_1}, \varphi_{\mathbf{x}_2}, \dots, \varphi_{\mathbf{x}_n}$ are linearly independent (for example $\varphi(r)=r^2$ in $V=\mathbb{R}$ is not a radial basis function).
    \item The kernels $\varphi_{\mathbf{x}_1}, \varphi_{\mathbf{x}_2}, \dots, \varphi_{\mathbf{x}_n}$ form a basis for a Haar Space, meaning that the interpolation matrix
    \begin{equation}\label{eq:1}
        \begin{bmatrix}
        \varphi(\|\mathbf{x}_1 - \mathbf{x}_1\|) & \varphi(\|\mathbf{x}_2 - \mathbf{x}_1\|) & \dots & \varphi(\|\mathbf{x}_n - \mathbf{x}_1\|) \\
        \varphi(\|\mathbf{x}_1 - \mathbf{x}_2\|) & \varphi(\|\mathbf{x}_2 - \mathbf{x}_2\|) & \dots & \varphi(\|\mathbf{x}_n - \mathbf{x}_2\|) \\
        \vdots & \vdots & \ddots & \vdots \\
        \varphi(\|\mathbf{x}_1 - \mathbf{x}_n\|) & \varphi(\|\mathbf{x}_2 - \mathbf{x}_n\|) & \dots & \varphi(\|\mathbf{x}_n - \mathbf{x}_n\|) \\
        \end{bmatrix}
    \end{equation}
    is non-singular \cite{Fasshauer2007} \cite{Wendland2005}.
\end{itemize}

Frequently utilized varieties of radial basis functions consist of:
\begin{itemize}
    \item \textbf{Gaussian RBF}:
    \[ \varphi(r) = \exp\left(-\frac{r^2}{2\sigma^2}\right) \]
    where $r$ is the distance between the input point and the center, and $\sigma$ is a parameter controlling the width of the Gaussian.
    
    \item \textbf{Multiquadric RBF}:
    \[ \varphi(r) = \sqrt{1 + \left(\frac{r}{\sigma}\right)^2} \]
    where $r$ is the distance between the input point and the center, and $\sigma$ is a parameter controlling the shape of the function.
    
    \item \textbf{Inverse Multiquadric RBF}: \label{rbf}
    \[ \varphi(r) = \frac{1}{\sqrt{1 + \left(\frac{r}{\sigma}\right)^2}} \]
    where $r$ is the distance between the input point and the center, and $\sigma$ is a parameter controlling the shape of the function.
    
    \item \textbf{Thin Plate Spline RBF}:
    \[ \varphi(r) = r^2 \log(r) \]
    where $r$ is the distance between the input point and the center.
\end{itemize}

Imagine a ball rolling around a landscape with hills and valleys. An attractor acts like the bottom of a valley. Regardless of where you place the ball on the landscape (starting conditions), if it rolls downhill long enough, it will eventually settle at the valley's bottom (the attractor). This signifies that the system (the ball) tends towards a specific set of values (the valley's position) over time. Thus, formally defining an attractor involves identifying a group of numeric values that a system naturally gravitates towards, irrespective of its initial parameters. 

Mathematical defintion of an attractor:\\
Let $t$ represent time and let $f(t,\cdot)$ be a function specifying the dynamics of the system. If $a$ is a point in an $n$-dimensional phase space, representing the initial state of the system, then $f(0,a)=a$, and for a positive value of $t$, $f(t,a)$ is the result of the evolution of this state after $t$ units of time. For example, if the system describes the evolution of a free particle in one dimension, then the phase space is the plane $\mathbb{R}^2$ with coordinates $(x,v)$, where $x$ is the position of the particle, $v$ is its velocity, $a=(x,v)$, and the evolution is given by

\[ f(t,(x,v)) = (x+tv, v). \]

An attractor is a subset $A$ of the phase space characterized by the following three conditions:

\begin{enumerate}
    \item $A$ is forward invariant under $f$: if $a$ is an element of $A$, then so is $f(t,a)$ for all $t > 0$.
    \item There exists a neighborhood of $A$, called the basin of attraction for $A$ and denoted $B(A)$, which consists of all points $b$ that "enter" $A$ in the limit $t \to \infty$. More formally, $B(A)$ is the set of all points $b$ in the phase space with the following property: For any open neighborhood $N$ of $A$, there is a positive constant $T$ such that $f(t,b) \in N$ for all real $t > T$.
    \item There is no proper (non-empty) subset of $A$ having the first two properties.
\end{enumerate}

Since the basin of attraction contains an open set containing $A$, every point that is sufficiently close to $A$ is attracted to $A$. The definition of an attractor uses a metric on the phase space, but the resulting notion usually depends only on the topology of the phase space  \cite{Milnor1985}. In the case of $\mathbb{R}^n$, the Euclidean norm \cite{celebi2010euclidean} is typically used, which is defined as \[
\| \textbf{x} \| = \sqrt{x_1^2 + x_2^2 + \dots + x_n^2}
\]. \label{euc}
Using these concepts we propose implementing a multi-branched radial basis neural network to help predict the chaotic and random behaviours of an attractor.

\section{Related Work}

Radial Basis networks have been extensively studied and proven effective in various classification tasks \cite{Wu2012Mar}\cite{leonard1991radial}. They offer a versatile framework for pattern recognition and data analysis, leveraging the flexibility of radial basis functions to model complex relationships within datasets. By capturing the intricate dynamics and nonlinear interactions inherent in real-world phenomena, Radial Basis networks contribute to advancing our understanding of complex systems and facilitating informed decision-making in fields ranging from communication systems \cite{jianping2002communication}\cite{yu2011advantages} to computational biology \cite{wouwer2004biological}\cite{ou2017identifying}.\\
While RBF layers offer valuable capabilities in certain modeling tasks, they alone may not be sufficient for capturing the rich dynamics and predicting chaotic and random behaviors in attractors. To address the complexities inherent in chaotic systems, more sophisticated and adaptable modeling approaches are required, which may involve combining RBF layers with other architectural components and techniques tailored to the specific characteristics of chaotic dynamics.\\
Attention mechanisms \cite{vaswani2023attention} have emerged as powerful tools in the realm of neural networks, offering sophisticated mechanisms for selectively focusing on relevant parts of input data while suppressing irrelevant information. Originally inspired by human cognitive processes, attention mechanisms have found widespread applications in various domains, including natural language processing, computer vision, and sequential data modeling.

\section{Dataset}

We use a pre existing kaggle dataset \cite{NIKITRICKY2023PhysicsAttractorTimeSeries}. This dataset comprises time series data originating from an unidentified physics attractor, synthesized through undisclosed governing rules. Manifesting intricate and chaotic dynamics, the attractor presents a challenge for analysis.\\
The dataset encompasses 36,700 data points, each delineating the positions of two points in a two-dimensional space at distinct time intervals. Collected over approximately 28 minutes, the dataset offers insights into the attractor's behavior over time. Notably, the system undergoes periodic resets, typically occurring upon reentry into a recurring loop. Table \ref{tab:dataset_description} shows the different variables in the dataset.\\
\renewcommand{\arraystretch}{1.2}
\begin{table}[h]
\centering
\begin{tabular}{lll}
\toprule
Variable & Type   & Definition \\
\midrule
time     & Float   & The time in seconds since the start of the simulation\\
distance  & Float   & Distance between both objects\\
angle1   & Angle   & Angle of the first object\\
pos1x    & Float   & X position of the first object\\
pos1y    & Float   & Y position of the first object\\
angle2   & Angle   & Angle of the second object\\
pos2x    & Float   & X position of the second object\\
pos2y    & Float   & Y position of the second object\\
\bottomrule
\end{tabular}
\caption{Description of variables present in the dataset.}
\label{tab:dataset_description}
\end{table}\\

\section{Methodology}

It defines the network architecture consisting of several components:
\\
\textbf{Branches}: Three separate branches are utilized, each focusing on learning the relationship between a specific pair of input columns.
    \begin{itemize}
        \item An RBFLayer: Performs the Radial Basis Function transformation on the input data; customizable parameters include the number of kernels $(K)$, output features $(F_{\text{o}})$, radial function $(\varphi)$, norm function $\|\cdot\|$, and normalization option ($\texttt{normalize}$). We use the \texttt{inverse\_multiquadric} function~\ref{rbf} and Euclidean norm~\ref{euc}.
        \item Dropout layer: Introduced with a probability of~$0.3$~to mitigate overfitting.
        \item AttentionLayer: Focuses on significant portions of the transformed data within the branch.
        \item Linear layers with $\operatorname{ReLU}(x) = \max\{0, x\}$
        and $\operatorname{tanh}(x) = \displaystyle\frac{e^{x}-e^{-x}}{e^{x}+e^{-x}}$
        activation functions for additional feature extraction and transformation.
    \end{itemize}
\textbf{Merging Layer}: Following the processing of each pair of columns within their respective branches, the outputs are concatenated. A linear layer with a ReLU activation function integrates the combined information.\\
\textbf{Output Layer}: A final linear layer with an output size of~$3$ projects the merged features onto the desired three-dimensional prediction.

Denote $x \in \mathbb{R}^3$ as the input vector having three features. Let ${\hat{y} \in \mathbb{R}^3}$ denote the output of the model. Each branch accepts a pair of input features represented as $x_i, x_j \in \{1, 2, 3\}$, satisfying ${i \neq j}$.

The forward function governs the data flow through the network:
\begin{itemize}
    \item Input Splitting: Separation of the input data $x$ into three distinct columns, representing the features: $x = (x_1, x_2, x_3)$.
    \item Branch Processing: Feeding each pair of columns into the assigned branch (branch1, branch2, or branch3); subsequently processed through their constituent layers, yielding an output per pair.
    \item Output Concatenation: The individual branch outputs, namely $(out_1, out_2, out_3)$, undergo concatenation along the feature dimension.
    \item Merging: Transmission of the concatenated outputs through the merging layer produces a unified representation.
    \item Prediction: Applying the merged features to the ultimate output layer yields the three-dimensional prediction $(\hat{y}_1, \hat{y}_2, \hat{y}_3)$.
\end{itemize}

\begin{figure}
    \centering
    \includegraphics[width=1\textwidth]{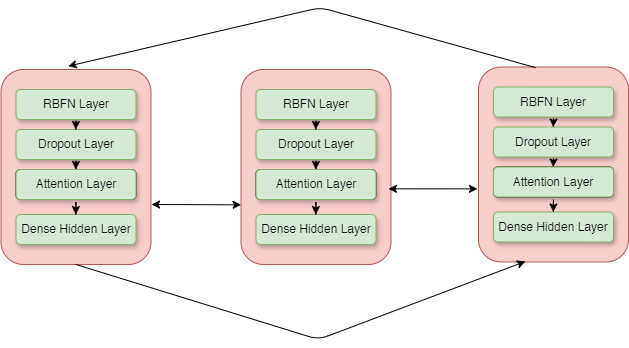}
    \caption{Proposed multi layer architecture}
    \label{model}
\end{figure}

\begin{figure}
    \centering
    \includegraphics[width=0.25\textwidth]{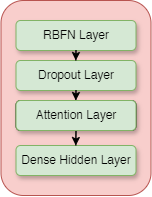}\label{single}
    \caption{Single Sequential Proposed Layer}
\end{figure}

This design enables the model to discern specific relationships amongst diverse input feature pairs while combining the learned features through the attention mechanism and merging stages for delivering the final prediction.

\section{Training}

We train the model on a singular NVIDIA A30 GPU. It takes 2 hours to train the model for \texttt{2000 epochs} with a \texttt{batch\_size=512}. We use the Mean Squared Error (MSE) loss function as our criterion:

\[
\mathrm{MSE}(\hat{\boldsymbol{y}}, \boldsymbol{y}) = \frac{1}{N} \sum_{i=1}^{N} \left(\hat{y}_{i} - y_{i}\right)^2
\]

Where $\hat{\boldsymbol{y}}$ represents the predicted values, $\boldsymbol{y}$ represents the actual target values, and $N$ is the total number of samples. The MSE computes the average of the squared differences between predicted and actual values, providing a measure of the model's performance in minimizing prediction errors.
We utilize \texttt{Adam} \cite{kingma2017adam} optimizer for our model:
\begin{align*}
m_t &= \beta_1 \cdot m_{t-1} + (1 - \beta_1) \cdot g_t, \\
v_t &= \beta_2 \cdot v_{t-1} + (1 - \beta_2) \cdot g_t^2, \\
\hat{m}_t &= \frac{m_t}{1 - \beta_1^t}, \\
\hat{v}_t &= \frac{v_t}{1 - \beta_2^t}, \\
\theta_{t+1} &= \theta_t - \frac{\eta}{\sqrt{\hat{v}_t} + \epsilon} \cdot \hat{m}_t,
\end{align*}
Where \( m_t \) and \( v_t \) are the first and second moment estimates, \( g_t \) is the gradient, \( \beta_1 \) and \( \beta_2 \) are the exponential decay rates for the moment estimates, \( \hat{m}_t \) and \( \hat{v}_t \) are bias-corrected estimates, \( \theta_t \) is the parameter at iteration \( t \), \( \eta \) is the learning rate, and \( \epsilon \) is a small constant to prevent division by zero.

Finally for comparison purposes we train with the same hyper parameters two models a singular branch \ref{single} and the proposed model \ref{model}. All implementations were done in PyTorch\cite{paszke2019pytorch}.

\section{Results}

\begin{figure}[!h]
    \centering
    \includegraphics[width=0.55\textwidth]{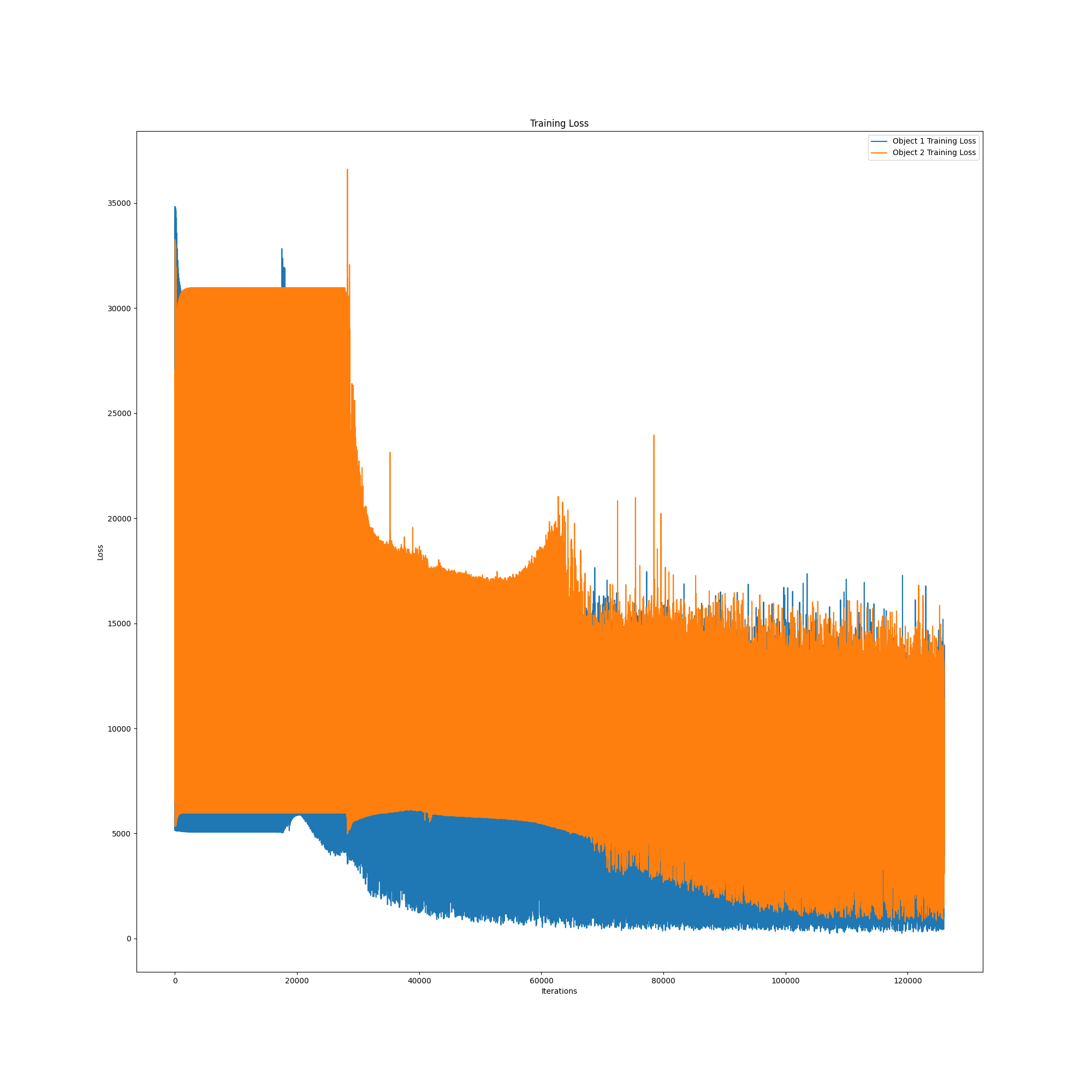}
    \caption{Loss over iterations of the Single Sequential Network}
    \label{ls}
\end{figure}

\begin{figure}[!h]
    \centering
    \includegraphics[width=0.55\textwidth]{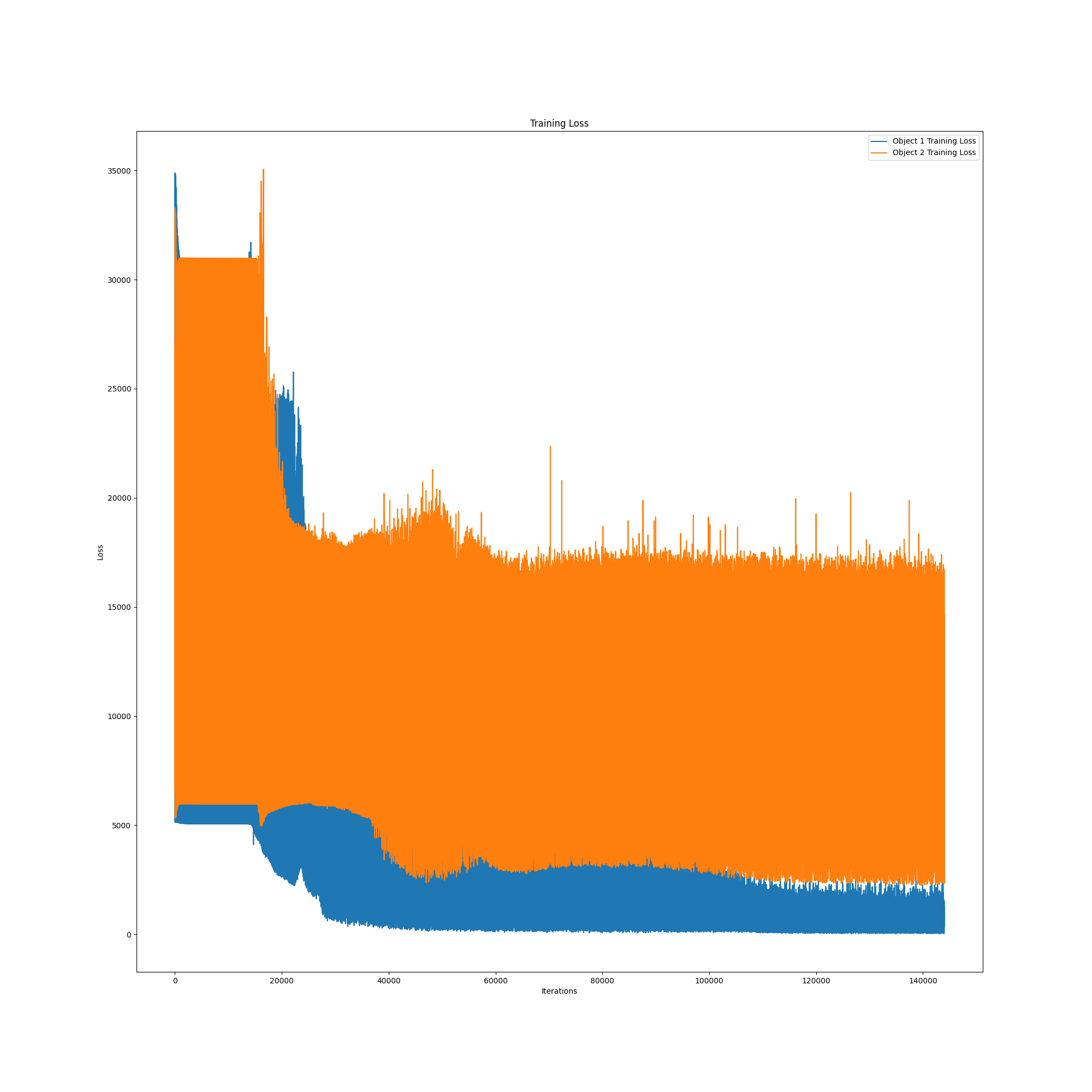}
    \caption{Loss over iterations of the Multi-Branched Network}
    \label{ms}
\end{figure}

\begin{itemize}
\item {Loss over iterations of the Single Sequential Network (Figure \ref{ls}})
The training loss for Object 1 (blue) starts high, sharply decreases, and then fluctuates around a lower level with some spikes. The training loss for Object 2 (orange) follows a similar pattern but maintains a higher overall loss throughout the training process. There are large spikes in the loss for both objects early in training, indicating potential instability or difficulty in the initial learning phase. The loss seems to stabilize and flatten out more towards the end of the training iterations shown.

\item {Loss over iterations of the Multi-Branched Network (Figure \ref{ms}})
has a  overall pattern is similar to Figure \ref{ls}, with Object 2's loss (orange) being consistently higher than Object 1's loss (blue).However, the initial large spikes in loss are more prominent and last longer compared to Image 1.The loss curves appear to flatten out and stabilize at a later point in the training process compared to Figure \ref{ls}.There are fewer small fluctuations and spikes in the loss curves once they stabilize, suggesting potentially smoother convergence.
\end{itemize}

In summary, while the overall trend of Object 2 having higher training loss is consistent across both images, the single sequential network exhibits more pronounced initial instability and takes longer to stabilize compared to the multi-branched architecture.

\begin{figure}[!h]
    \centering
    \includegraphics[width=0.55\textwidth]{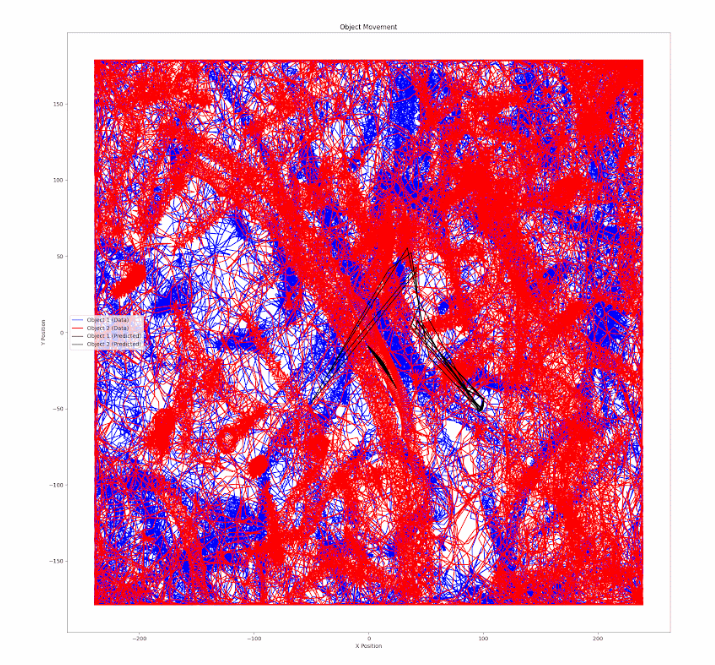}
    \caption{Output of the single sequential layer}
    \label{sout}
\end{figure}

\begin{figure}[!h]
    \centering
    \includegraphics[width=0.55\textwidth]{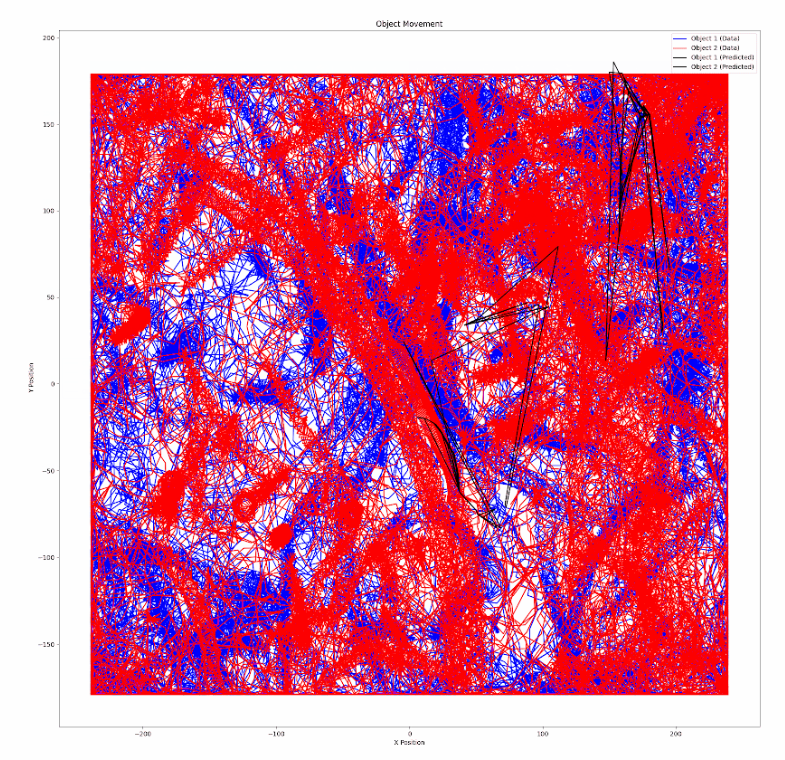}
    \caption{Output of the multi layer}
    \label{mout}
\end{figure}

We next compare the outputs of the single sequential layer and the multi layered architecture. Figure \ref{sout} shows the object movement for the single sequential layer and Figure \ref{mout} shows the object movement for the multi layered architecture.\\

The predicted paths (black lines) for the single sequential layer \ref{sout} are relatively centralized and seem to capture some linear segments of the trajectories. The overall pattern shows dense and tangled paths, which is typical of chaotic systems. The black lines appear to follow the chaotic nature to some extent but might be too centralized and not dispersed enough to fully capture the randomness.
The predicted paths (black lines) for the multi layer architecture layer \ref{mout} are also centralized but show slight shifts compared to the output of the single sequential layer.
This output also has dense and tangled paths, consistent with chaotic behavior. The black lines appear to capture more variability and slight shifts, which might better reflect the unpredictability of chaotic systems.

\clearpage

\section{Conclusion}
In conclusion, this paper has explored the application of Radial Basis Function Neural Networks (RBFNNs) in predicting chaotic and random behaviors. Through a comprehensive review of related work, we have highlighted the strengths and limitations of RBFNNs in capturing the complex dynamics of chaotic systems. Leveraging insights from chaos theory and neural network architecture, we have proposed novel approaches for enhancing the predictive capabilities of RBFNNs with attention mechanisms. \\
Our results demonstrate the effectiveness of our proposed methods in predicting chaotic and random behaviors. A comparison of object movement predictions illustrated in our visual results indicates that our enhanced RBFNN model effectively captures the inherent variability and unpredictability of chaotic systems. Specifically, in Figure \ref{mout} prediction paths exhibited greater variability and subtle shifts, closely aligning with the expected characteristics of chaotic behavior. This confirms that our model can realistically reflect the randomness and sensitivity to initial conditions typical of chaotic systems.\\
Overall, this paper contributes to advancing our understanding of chaotic systems and lays the groundwork for future research in utilizing RBFNNs for predictive modeling in complex dynamical systems.

\section{Limitations}

Chaotic systems often require ongoing monitoring and adjustments to plans.  Since small changes can have significant impacts, staying updated on the current state of the system is crucial.
We understand that chaos can not be truly predicted and understood.

\section{Reproducibility}

Results can be reproduced from the code present in my \href{https://github.com/chungimungi/Attractor-Chaos}{GitHub} repository.

\section{Acknowledgement}
We acknowledge the work of Alessio Russo who originally implemented RBFNNs in PyTorch. His work is available on his GitHub\cite{pythonvrft}.

\bibliographystyle{unsrt}  
\bibliography{references}  

\begin{thebibliography}{10}

\bibitem{rbf}
{Contributors to Wikimedia projects}.
\newblock {Radial basis function - Wikipedia}, 2024.

\bibitem{Fasshauer2007}
Gregory~E. Fasshauer.
\newblock {\em {Meshfree Approximation Methods with MATLAB}}.
\newblock World Scientific Publishing Co. Pte. Ltd., Singapore, 2007.

\bibitem{Wendland2005}
Holger Wendland.
\newblock {\em {Scattered Data Approximation}}.
\newblock Cambridge University Press, Cambridge, 2005.

\bibitem{Milnor1985}
John Milnor.
\newblock On the concept of attractor.
\newblock {\em Communications in Mathematical Physics}, 99(2):177--195, Jun 1985.

\bibitem{celebi2010euclidean}
M.~Emre Celebi, Fatih Celiker, and Hassan~A. Kingravi.
\newblock On euclidean norm approximations, 2010.

\bibitem{Wu2012Mar}
Yue Wu, Hui Wang, Biaobiao Zhang, and K.-L. Du.
\newblock {Using Radial Basis Function Networks for Function Approximation and Classification}.
\newblock {\em International Scholarly Research Notices}, 2012, March 2012.

\bibitem{leonard1991radial}
James~A Leonard and Mark~A Kramer.
\newblock Radial basis function networks for classifying process faults.
\newblock {\em IEEE Control Systems Magazine}, 11(3):31--38, 1991.

\bibitem{jianping2002communication}
Deng Jianping, Narasimhan Sundararajan, and P~Saratchandran.
\newblock Communication channel equalization using complex-valued minimal radial basis function neural networks.
\newblock {\em IEEE Transactions on neural networks}, 13(3):687--696, 2002.

\bibitem{yu2011advantages}
Hao Yu, Tiantian Xie, Stanis{\l}aw Paszczynski, and Bogdan~M Wilamowski.
\newblock Advantages of radial basis function networks for dynamic system design.
\newblock {\em IEEE Transactions on Industrial Electronics}, 58(12):5438--5450, 2011.

\bibitem{wouwer2004biological}
A~Vande Wouwer, Christine Renotte, and Ph~Bogaerts.
\newblock Biological reaction modeling using radial basis function networks.
\newblock {\em Computers \& chemical engineering}, 28(11):2157--2164, 2004.

\bibitem{ou2017identifying}
Yu-Yen Ou et~al.
\newblock Identifying the molecular functions of electron transport proteins using radial basis function networks and biochemical properties.
\newblock {\em Journal of Molecular Graphics and Modelling}, 73:166--178, 2017.

\bibitem{vaswani2023attention}
Ashish Vaswani, Noam Shazeer, Niki Parmar, Jakob Uszkoreit, Llion Jones, Aidan~N. Gomez, Lukasz Kaiser, and Illia Polosukhin.
\newblock Attention is all you need, 2023.

\bibitem{NIKITRICKY2023PhysicsAttractorTimeSeries}
NIKITRICKY.
\newblock Physics attractor time series dataset, 2023.

\bibitem{kingma2017adam}
Diederik~P. Kingma and Jimmy Ba.
\newblock Adam: A method for stochastic optimization, 2017.

\bibitem{paszke2019pytorch}
Adam Paszke, Sam Gross, Francisco Massa, Adam Lerer, James Bradbury, Gregory Chanan, Trevor Killeen, Zeming Lin, Natalia Gimelshein, Luca Antiga, Alban Desmaison, Andreas Köpf, Edward Yang, Zach DeVito, Martin Raison, Alykhan Tejani, Sasank Chilamkurthy, Benoit Steiner, Lu~Fang, Junjie Bai, and Soumith Chintala.
\newblock Pytorch: An imperative style, high-performance deep learning library, 2019.

\bibitem{pythonvrft}
Alessio Russo.
\newblock Pytorch rbf layer, 2021.

\end{thebibliography}

\end{document}